\documentclass[final,1p,times]{elsarticle}

\usepackage{amssymb}
\usepackage{amsmath}

\usepackage{lineno}

\usepackage{dsfont}
\usepackage{algorithmic}
\usepackage{algorithm}
\usepackage{array}
\usepackage{hyperref}
\usepackage{xurl}
\usepackage{textcomp}
\usepackage{stfloats}
\usepackage{hyperref}       
\usepackage{url}            
\usepackage{booktabs}       
\usepackage{nicefrac}       
\usepackage{microtype}      
\usepackage{xcolor}         
\usepackage{graphicx}
\usepackage{cite}
\usepackage{multirow}
\usepackage{xcolor}
\usepackage{threeparttable}
\usepackage[capitalise]{cleveref}
\usepackage{tikz}
\usepackage{siunitx}

\journal{}

\begin{document}

\begin{frontmatter}



\title{\LARGE{Managing Geological Uncertainty in Critical Mineral Supply Chains: A POMDP Approach with Application to U.S. Lithium Resources}}

\author[1]{Mansur   Arief}
\affiliation[1]{organization={Aeronautics \& Astronautics, Stanford University}, city={Stanford}, state={CA}, country={USA}}

\author[2]{Yasmine Alonso}
\affiliation[2]{organization={Computer Science, Stanford University}, city={Stanford}, state={CA}, country={USA}}

\author[3]{CJ Oshiro}
\affiliation[3]{organization={Computer Science, Santa Clara University}, city={Santa Clara}, state={CA}, country={USA}}

\author[4]{William Xu}
\affiliation[4]{organization={Materials Science \& Engineering, Stanford University}, city={Stanford}, state={CA}, country={USA}}

\author[1]{Anthony Corso}

\author[5]{David Zhen Yin}
\affiliation[5]{organization={Earth \& Planetary Sciences, Stanford University}, city={Stanford}, state={CA}, country={USA}}

\author[5]{Jef K. Caers}

\author[1]{Mykel J. Kochenderfer}




\begin{abstract}

The world is entering an unprecedented period of critical mineral demand, driven by the global transition to renewable energy technologies and electric vehicles. This transition presents unique challenges in mineral resource development, particularly due to geological uncertainty—a key characteristic that traditional supply chain optimization approaches do not adequately address. To tackle this challenge, we propose a novel application of Partially Observable Markov Decision Processes (POMDPs) that optimizes critical mineral sourcing decisions while explicitly accounting for the dynamic nature of geological uncertainty. Through a case study of the U.S. lithium supply chain, we demonstrate that POMDP-based policies achieve superior outcomes compared to traditional approaches, especially when initial reserve estimates are imperfect. Our framework provides quantitative insights for balancing domestic resource development with international supply diversification, offering policymakers a systematic approach to strategic decision-making in critical mineral supply chains. 

\end{abstract}







\end{frontmatter}



%
%
%
%
%

\section{Introduction}

The world is embarking on an unprecedented period of critical mineral demand, marking a unique chapter in human history that has no historical parallel~\citep{gielen2021critical, ambrose2020understanding}. Unlike previous industrial transformations, which relied primarily on widely available resources like iron and coal, the rapid shift toward renewable energy technologies and electric vehicles has created extraordinary challenges in securing reliable supplies of critical minerals, particularly lithium~\citep{mohr2012lithium, tadesse2019beneficiation, kamran2023critical}. Recent studies highlight that the projected demand for lithium by 2040 could exceed all historical lithium production combined~\citep{kalair2021role, kushnir2012time}, creating supply challenges that cannot be addressed with traditional industrial scaling approaches.

These unprecedented demands are accompanied by significant uncertainties that distinguish critical mineral supply chains from traditional supply chain studies (e.g., \citep{yu2024structural, wang2023energy}), which stem from three primary sources: geological uncertainty in resource estimation, economic and social uncertainty in market dynamics, and geopolitical uncertainty in international relations~\citep{graham2021lithium, christmann2015global}. The economic and environmental implications of rapidly scaling lithium production add another layer of complexity, with studies showing significant risks in investment as well as extensive ecological footprints from water-intensive extraction processes~\citep{kaunda2020potential}. The challenge of obtaining social licenses for mining projects adds more complexity to the development timeline~\citep{komnitsas2020social}. Furthermore, the geographic concentration of lithium resources (with the majority of known reserves to date located in a limited number of countries, including Bolivia, Chile, Argentina, China, USA, and Australia), further compounds these challenges~\citep{christmann2015global, lunde2020lessons, ebensperger2005lithium}. This concentration raises concerns about overreliance on specific countries, which may be susceptible to political and economic instabilities~\citep{hailes2022lithium, sanchez2023geopolitics, tian2021features}.

\begin{figure}[h]
    \centering
    \includegraphics[width=0.8\textwidth]{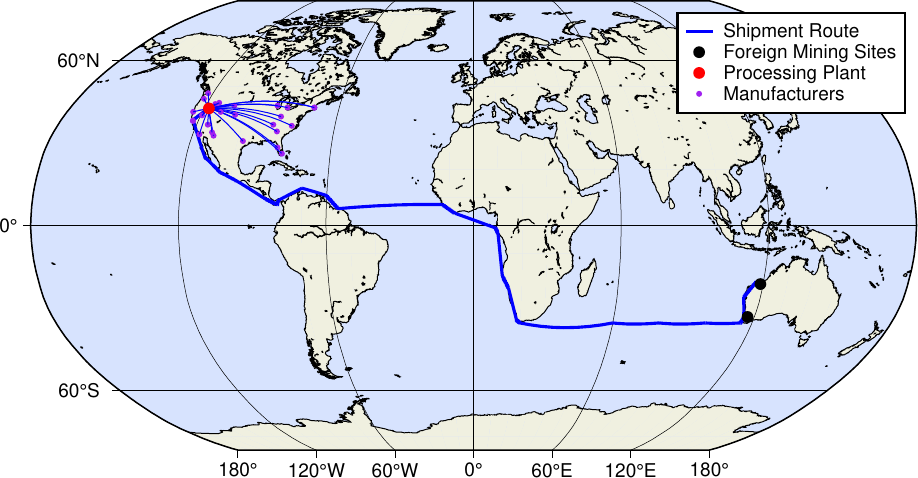}
    \caption{Lithium supply chain example from foreign mining sites to processing plant and manufacturing facilities in the U.S. The lithium mineral is transported by sea from Australia to the U.S. West Coast, then by rail to Nevada. The route covers at least 17,062 nautical miles and takes approximately 10 weeks on average~\citep{ports_searoute}.}
    \label{fig:supply_chain}
\end{figure}

The U.S. lithium supply chain exemplifies these challenges. As illustrated in~\autoref{fig:supply_chain}, the current supply chain relies heavily on importing lithium-bearing minerals from geographically distant locations such as Australia, which involves complex maritime routes spanning more than 17,062 nautical miles and requiring approximately 10 weeks of transportation time~\citep{ports_searoute}. The route includes sea transportation from Australian mining sites to U.S. West Coast ports, followed by rail transportation to processing facilities in Nevada, where several battery manufacturers are located. This extended supply chain exposes vulnerabilities to various disruptions and contributes to both economic and environmental costs. While substantial domestic lithium reserves exist (such as the Thacker Pass deposit in Nevada with approximately 1.56 MMt of lithium supply), the exact volume and quality of these reserves remain uncertain~\citep{uji2023pursuing}. This uncertainty, combined with environmental and social concerns about land use, water consumption, and carbon emissions, has created significant challenges for the development of domestic resources. The mining industry cannot ignore these considerations, as they could result in substantial social and economic penalties.

While these economic, social, and geopolitical uncertainties pose nontrivial challenges to critical mineral supply chains, the inability to accurately assess subsurface mineral deposits without extensive exploration represents a distinctive characteristic of mineral resource development. Traditional optimization approaches to supply chain design typically rely on accurate estimates of resource availability and treat uncertainty around these estimates through scenario analysis or stochastic programming \citep{chen2023uncertainty}. However, these methods fail to capture the dynamic nature of geological uncertainty and its evolution through exploration and extraction activities~\citep{granholm2021national}. More importantly, they do not account for the value of information that can be gained through systematic exploration and development activities \citep{caers2022efficacy}. This limitation suggests the need for more sophisticated modeling approaches that can adapt to new information and update resource estimates over time.

We propose an intelligent agent-based approach, specifically one based on a model known as a Partially Observable Markov Decision Process (POMDP)~\citep{kochenderfer2022algorithms}, to account for the dynamic nature of geological uncertainty. We believe that incorporating belief updates within the planning process plays a crucial role for critical mineral supply chains. Our hypothesis is based on two key observations. First, POMDPs explicitly model the reduction of uncertainty over time through belief updates, a crucial aspect of mineral resource development that is often overlooked in static optimization approaches. Second, the sequential nature of POMDP decision-making aligns naturally with the gradual, iterative process of resource exploration and development~\citep{williams2011resolving, hansen2016partially}. The significance of our approach extends beyond pure optimization metrics to inform robust strategic policy development for critical mineral supply chains that considers geological uncertainty~\citep{csis2024friendshoring}.

Our contributions can be summarized as follows. First, we demonstrate that POMDP-based policies can achieve superior outcomes, especially when initial reserve estimates are imperfect, highlighting the value of sequential decision-making over traditional single-shot optimization approaches. Second, we show how updated beliefs provide ways to quantify geological uncertainty and can lead to more robust resource development strategies by trading off exploration costs with the risk of imperfect initial estimates. Third, using a simplified U.S. lithium supply chain case study, we demonstrate how our approach can help balance the complex tradeoffs involved in developing domestic resources and maintaining diverse international supply sources, a general challenge that can easily be further adopted for other critical minerals to inform policymaking in future studies.


\section{Related Work}\label{sec:literature_review}

This section reviews literature on uncertainty characterization in mineral supply chains and sequential decision-making frameworks, with emphasis on how existing approaches address geological uncertainty and information dynamics in resource development.

\subsection{Uncertainty in Mineral Supply Chains}

Critical mineral supply chains face three primary sources of uncertainty that distinguish them from traditional industrial supply chains \citep{caers2022efficacy, mcnulty2021barriers}. The first and most distinctive is geological uncertainty, which fundamentally shapes the decision-making landscape in ways unique to mineral supply chains. This uncertainty stems from the inherent difficulty in accurately assessing subsurface mineral deposits~\citep{mern2023intelligent}, and unlike manufactured goods where production capacity can be estimated with relative certainty, mineral resources remain partially unknown until significant investment in exploration and extraction has occurred.

Geological uncertainty gives rise to several interconnected forms of uncertainty that compound the complexity of resource assessment and development planning \citep{caers2022efficacy}. Resource quantity uncertainty affects our understanding of recoverable mineral volumes, while quality uncertainty influences the economic viability of extraction through variations in grade and composition. Spatial uncertainty impacts the distribution and accessibility of resources, affecting development costs and extraction strategies. Recovery uncertainty further complicates planning by introducing variability in extraction efficiency and processing yields.

Economic uncertainty represents the second major challenge, encompassing market price volatility, development costs, and technological change~\citep{hailes2022lithium}. The economic viability of lithium projects is particularly sensitive to these uncertainties because there are high upfront costs of mine development and long lead times between investment and production. The third source, geopolitical uncertainty, includes trade policies, environmental regulations, and international relations~\citep{sanchez2023geopolitics, tian2021features}. While these latter uncertainties are common to many supply chains, their interaction with geological uncertainty creates unique challenges in the critical minerals sector.

In this work, we focus exclusively on geological uncertainty, as it represents the most distinctive characteristic of mineral supply chains and, crucially, the only uncertainty that can be systematically reduced through exploration and information gathering. Unlike economic and geopolitical uncertainties, which are largely external to the decision-making process, geological uncertainty can be actively managed through sequential exploration and development decisions. An earlier work toward this direction is explored in~\citep{armstrong2021adaptive}, implementing a rolling horizon optimization with Bayesian updating. While it is more computationally simpler, it may not fully capture the value of information gathering actions since it doesn't explicitly model how future beliefs will change. In contrast, we propose a POMDP-based approach explicitly model both the state uncertainty and how actions affect future observations and beliefs. The solution gives a policy that maps beliefs to actions, accounting for both immediate rewards and information gathering value, allowing us to clearly demonstrate the value of sequential decision-making in mineral supply chains and gaining deeper insights for strategic policy making.

\subsection{Sequential Decision-Making for Mineral Supply Chains}

The limitations of traditional optimization approaches have led to growing interest in sequential decision-making frameworks, particularly Partially Observable Markov Decision Processes (POMDPs). Recent work has demonstrated the effectiveness of POMDPs in subsurface exploration and planning~\citep{mern2023intelligent, corso2024sequentially}. While exact solutions to POMDPs are computationally challenging for real-world problems~\citep{shani2013survey}, approximate methods have shown promising results. These include point-based approaches focusing on reachable belief states, Monte Carlo tree search variants for continuous state spaces~\citep{gelly2011monte}, and online planning methods that interleave planning and execution~\citep{silver2010monte}. 

\subsection{Research Gap and Our Contribution}
Despite extensive research on mineral exploration and decision-making under uncertainty, significant gaps remain in approaches that explicitly model and leverage the sequential nature of geological uncertainty reduction in strategic supply chain planning. Current methods typically treat geological uncertainty as static, failing to capture both the value of exploration activities and the adaptive nature of mineral development decisions. While existing POMDP applications often focus on operational exploration decisions, our study addresses strategic supply chain sourcing decisions \citep{csis2024friendshoring, mcnulty2021barriers}, demonstrating how technical uncertainties, particularly geological ones, significantly impact strategic planning. By applying POMDPs to this context, we advance both the theoretical understanding of supply chain optimization under uncertainty and the development of practical, robust strategies for critical mineral resource management.

\section{POMDP Formulation}\label{sec:formulation}

A POMDP is defined by a tuple $\langle S, A, O, T, R, Z, \gamma \rangle$, where $S$ is the state space, $A$ is the state space, $O$ is the observation space, $T$ is the transition function, $R$ is the reward function, $Z$ is the observation function, and $\gamma$ is the discount factor. At each time step $t$, the agent selects an action $a_t \in A$ based on the history, receives an observation $o_t \in O$ and a reward $r_t$ based on the action taken $a_t$, and transitions from the current state $s_t \in S$ to the next state $s_{t+1} \in S$. Uncertainties in the problem are modeled by the observation function $Z$ and the transition function $T$. In this case, observation $o_t$ is sampled from distribution $Z( \cdot \mid a_t, s_{t+1})$, and the next state $s_{t+1}$ is sampled from distribution $T( \cdot \mid s_t, a_t)$. 

With this setup, the history of the agent's interactions with the environment is often represented as a belief $b_t \in B$, which is a probability distribution over the state space. A POMDP policy $\pi$ is a mapping from beliefs to actions, $\pi: B \rightarrow A$, where $B$ is the belief space. The goal of POMDP solvers is to find an optimal policy $\pi^*$ that maximizes the expected cumulative reward over time. A crucial aspect of the POMDP framework is describing its main components that are specific to the problem at hand. In the subsequent sections, we will describe the POMDP formulation for the lithium sourcing problem, including the state space, action space, observation space, transition function, reward function, and belief update mechanism.

\subsection{States}

Assuming we consider $n$ lithium deposit sites (some of which are domestic and some of which are foreign), the state in the POMDP model represents the possible configurations of the lithium supply chain, including:
\begin{itemize}
\item the true \underline{v}olume of lithium carbonate equivalent (LCE) reserves at each site: $[v_1, v_2, \ldots, v_n]^\intercal$, 
\item the indicators for whether \underline{m}ining operation is in progress at each site: $[m_1, m_2, \ldots, m_n]^\intercal$, 
\item the amount of LCE \underline{i}mported from foreign sources $i$, 
\item the amount of LCE extracted from \underline{d}omestic sources $d$, and 
\item the current \underline{t}ime step $t$. 
\end{itemize}
A POMDP state $s$ is therefore defined as
\begin{equation}
    s = \left( 
            \begin{bmatrix}
                v_1 \\
                v_2 \\
                \vdots \\
                v_n \\
            \end{bmatrix},
            \begin{bmatrix}
                m_1 \\
                m_2 \\
                \vdots \\
                m_n \\
            \end{bmatrix},
            i, d, t 
        \right),
\end{equation}
and the state space $S$ is the Cartesian product of the possible values of each of these variables. 

To make the problem more computationally tractable, we discretize the volume of LCE reserves at each site into finite intervals. We also discretize the LCE imported $i$, LCE extracted domestically $d$, and time step $t$. Despite this discretization, given the high-dimensionality of the state space, the problem is highly non-trivial to solve.

\subsection{Actions}

At each time step, the agent can choose to \textsc{explore}, \textsc{build}, or \textsc{restore} at any chosen deposit site $j$, or \textsc{do~nothing}.

\begin{itemize}
    \item \textsc{explore}: The agent conducts exploration and collects measurements to estimate the volume of LCE reserves $v_j$ at chosen site $j$. The measurements obtained from this action are noisy and uncertain, reflecting the challenges of estimating the true volume of LCE reserves from geological surveys and exploration activities. When taking \textsc{explore} action, the agent incurs an \underline{e}xploration cost $c_e$ and receives a noisy observation.
    \item \textsc{build}: The agent invests in building infrastructure and equipment for mining operations at chosen site $j$. This action incurs a \underline{b}uild cost $c_b$ and starts operating at the next time step. A crucial aspect of this action is that it allows the agent to start mining at site $j$ at the next time step and will only terminate when the agent takes the \textsc{restore} action.
    \item \textsc{restore}: The agent invests in rehabilitation and restoration activities at chosen site $j$. This action incurs a \underline{r}estoration cost $c_r$ representing post-mining activity to restore the site after mining operations have ceased. As mentioned before, the agent will only take this action when the mine is operational, after which it will cease to operate at site $j$.
    \item \textsc{do~nothing}: The agent continues to operate established mines without exploring, building, or restoring at any of the mine sites. This action does not incur any additional extra cost, other than the cost of operating and maintaining the already established mines.
\end{itemize}

\noindent The action space is therefore 
\begin{align} \nonumber
A =\{&\textsc{do~nothing}\} \bigcup_{j=1}^n \{\textsc{explore}_j, \textsc{build}_j, \textsc{restore}_j\}.
\end{align}
The agent can choose to take any action $a \in A$, subject to the constraints of the problem. For instance, it only makes sense to \textsc{explore} at a site where the mine has not been built yet or to \textsc{restore} after the mine has been built.

\subsection{Observations}

The observation that the agent obtains at a given time step depends on the action taken and the true state of the environment at that time step. To operationalize this, the observation is represented as a vector of measurements from the $n$ sites, where the lack of a measurement is represented by a value of $-1$. If the agent takes the \textsc{explore} action at site $j$, it receives a noisy measurement of the true volume of LCE reserves $v_j$ at site $j$, modeled as a Gaussian sample $\tilde v_j \sim N(v_j, \sigma_{o})$, where $\sigma_{o}$ is the observation noise representing the uncertainty in the measurement process. In such a case, $o$ is a vector of $-1$ values except for the site $j$ where the value is $\tilde v_j$. Meanwhile, if the agent takes the \textsc{build} or \textsc{restore} action, the agent receives no observation, hence $o$ is simply a vector of $-1$ values. We also discretize the observation space to make the problem computationally tractable.

\subsection{State Transitions}

The state transition uncertainty modeled in this work focuses on the stochastic nature of mining operations and disruptions in long-distance transportation. At each time step, the LCE volume extracted from each site $j$, $E_j$, is a discrete random variable that follows distribution $\phi_j$, constrained on the remaining reserve volume $v_j$. If the source is a foreign site, there is an added layer of uncertainty due to potential disruptions in transportation. Such disruptions might cause some volume loss $L_j$, which is modeled as a discrete random variable that follows distribution $\psi_j$. The total volume of LCE extracted from foreign site $j$ that reaches the processing plant is $Z_j = E_j - L_j$. 

At the next time step, the state is updated as follows:
\begin{itemize}
    \item the remaining volume of LCE reserves is updated to be $v_j' = v_j - E_j$ if the mine is operational, or, $v_j' = v_j$, if the mine is not operational,
    
    \item the mining operations at each site are updated to be $m_j' = 1$ if the agent takes the \textsc{build} action at site $j$, $m_j' = 0$ if the agent takes the \textsc{restore} action at site $j$, or $m_j' = m_j$ if the agent takes any other action,

    \item the LCE imported becomes $i' = i + \sum_{j \in J_f}^n Z_j$, where $J_f$ is the set of foreign sites,

    \item the LCE extracted from domestic sources becomes $d' = d + \sum_{j \in J_d}^n E_j$, where $J_d$ are the set of domestic sites, and finally,
    
    \item the time step is incremented by one, $t' = t + 1$.
\end{itemize}

\subsection{Rewards}

We incorporate the economic, environmental, and social aspects of the problem into the reward function $R$, defined as the weighted sum of the following components:

\begin{itemize}
    \item \textbf{Domestic Mining Penalty ($R_1$).} The agent is penalized an amount $p_d$ when building \underline{d}omestic mining infrastructure before a certain time delay goal $t_d$, reflecting the challenges of obtaining social license and environmental permits for mining operations:
    \begin{equation}
        R_1(s, a) = \left\{
            \begin{aligned}
                -&p_d,~\text{if } t < t_d \text{ and } a \in \bigcup_{j=1}^n \{ \textsc{build}_j \} \\
                &0,~\text{ otherwise}
            \end{aligned} \right. ,
    \end{equation}
    where $t$ is the current time of state $s$.

    \item \textbf{CO$_2$ Emission Penalty ($R_2$).} The agent is penalized for the amount of CO$_2$ emissions generated by the mining and transportation operations:
    \begin{equation}
        R_2(s, a) = \sum_{j=1}^n E_j e_j + \mathds{1} \left( a = \textsc{restore}_j \right)  r_j,
    \end{equation}
    where $e_j$ is the CO$_2$ \underline{e}mission factor for both mining at site $j$ and transporting lithium from site $j$ to the processing plant, which depends on the type of lithium deposit, distance from the processing plant, and transportation mode and $r_j$ is the CO$_2$ absorption factor for restoration project at site $j$.

    \item \textbf{Demand Unfulfilled Penalty ($R_3$).} The agent is penalized for the amount of unfulfilled demand at each time step:
    \begin{equation}
        R_3(s, a) = \max(0, d_t - \rho l_t),
    \end{equation}
    where $d_t$ is the \underline{d}emand for lithium at time $t$, $\rho$ the extraction factor during processing at the plant, and 
    \begin{equation}
        l_t = \sum_{j \in J_d} E_j + \sum_{j \in J_f} Z_j
    \end{equation}
    is the amount of \underline{l}ithium-bearing mineral before processing.

    \item \textbf{Profit ($R_4$).} The profit is calculated as the revenue from selling the processed lithium minus the total costs of exploration, building/investment, mining/operations, restoration, transportation and processing. The revenue is calculated as:
    \begin{equation}
        \text{revenue}(s,a) = \min(d_t, \rho l_t) \times  p_{Li},
    \end{equation}
    where $p_{Li}$ is the market \underline{p}rice of lithium. The total costs are calculated as:
    \begin{align}
        \text{cost}(s,a) &= \underbrace{c_e \times \mathds{1} \left(a \in \bigcup_{j=1}^n \{\textsc{explore}_j\} \right)}_{\text{exploration cost}} + \underbrace{c_b \times \mathds{1} \left(a \in \bigcup_{j=1}^n \{\textsc{build}_j\} \right)}_{\text{build/investment cost}} \nonumber \\
        &+ \underbrace{c_o \times \sum_{j=1}^n \mathds{1} \left(m_j = 1 \right)}_{\text{mining/operations cost}} \nonumber + \underbrace{c_r \times \mathds{1}\left(a \in \bigcup_{j=1}^n \{\textsc{restore}_j\} \right)}_{\text{restoration cost}} \nonumber \\
        & + \underbrace{(c_t + c_p) \times l_t}_{\text{transport and processing cost}},
    \end{align}
    where $\mathds{1}(\cdot)$ is the indicator function, and $c_t$ and $c_p$ are the  \underline{t}ransportation and  \underline{p}rocessing costs, respectively. The profit is then calculated as
    \begin{equation}
        R_4(s,a) = \text{revenue}(s,a) - \text{cost}(s,a).
    \end{equation}

\end{itemize}
Finally, the total reward function is calculated as the weighted sum of the individual components:
\begin{equation}
    R(s,a) = \sum_{i=1}^4 w_i R_i(s,a),
\end{equation}
where $w_1, w_2, w_3, w_4$ are the weights that determine the relative importance of each objective in the reward function. These weights can be adjusted based on the strategic priorities of the decision maker.

\subsection{Solving the POMDP}

Given the high-dimensionality of the state space, we use online POMDP solvers to maximize the expected cumulative reward over time, formulated as
\begin{equation}
    \max_{\pi} \mathbb{E} \left[ \sum_{t=0}^{\infty} \gamma^t R(s_t, \pi(b_t)) \right],
\end{equation}
where the expectation is taken over the distribution of the initial state, and the distribution of the transition and observation functions that make up the belief update mechanism. Here, $\gamma \in (0, 1)$ is the discount factor, which determines the relative importance of short-term versus long-term rewards. This factor is crucial for problems that deal with long-term planning, such as the lithium sourcing problem. Furthermore, $\pi(b_t)$ is the policy that maps the belief $b_t$ to an action $a_t$. Here, a belief $b_t$ is a probability distribution over the state space, representing the agent's uncertainty about the true state of the environment at time $t$. For our problem, this is mainly due to the uncertainty in the geological reserves at each site, which is modeled as a Gaussian distribution. At each time step, the agent updates its belief with the Kalman filter update equations using the noisy observation $o_t$ corresponding to the action $a_t$ taken at time $t$.

\subsection{Belief Update}

The belief update is performed at each time step to maintain an accurate representation of the agent's uncertainty about the true state of the environment. Given a prior belief $b$ and an observation $o$, the agent updates its belief to obtain the posterior belief $b'$. However, a meaningful belief update only occurs when the agent takes the \textsc{explore} action, as measurements are only available in this case. In other cases, the belief remains unchanged.

In this model, we use a Gaussian distribution as the belief representation. For each site $j$, the prior belief $b$ is $N(\mu_j, \sigma_j^2)$. Given that the noisy measurement $\tilde v_j$ is also a Gaussian random variable, the posterior belief $b'$ is also a Gaussian distribution with mean and variance given by the Kalman filter update equations. At each Kalman update, the mean and variance of the posterior belief are updated as follows~\citep{kochenderfer2022algorithms}:
\begin{align}
    \mu_j' &= \mu_j + K_j(\tilde v_j - \mu_j), \\
    \sigma_j'^2 &= (1 - K_j) \sigma_j^2,
\end{align}
where $K_j$ is the Kalman gain, calculated as
\begin{equation}
    K_j = \frac{\sigma_j^2}{\sigma_j^2 + \sigma_o^2}.
\end{equation}
Here, $\sigma_o$ is the observation noise representing the uncertainty in the measurements process. The posterior belief for site $j$ after \textsc{explore}$_j$ action is therefore $b' = N(\mu', \sigma'^2)$. The belief update mechanism allows the agent to maintain an accurate representation of the true state of the environment and make informed decisions based on the available information.

Arguably, the belief update mechanism is the most valuable aspect of the POMDP framework to address geological uncertainty in mineral exploration. It enables the agent to adjust the exploration strategy based on the available information and make informed decisions about where to explore next. In contrast, classical optimization approaches that rely fully on the initial estimate of the reserves are not able to adjust the exploration strategy on the fly, and are therefore not able to take advantage of the information obtained from the exploration actions. We will show via numerical experiments that this is crucial to achieve a good performance for long-term critical mineral supply chain planning.

\section{Numerical Experiments}\label{sec:experiments}

The goal of our experiment is to evaluate the performance of the POMDP framework against various benchmarks in optimizing lithium sourcing decisions under uncertainty. The benchmarks include:
\begin{itemize}
    \item \textsc{random} policy, which takes actions randomly without considering the state of the environment,
    \item \textsc{greedy} heuristic, which takes actions based on an initial estimate of the reserves and no exploration, aiming to maximize the amount of LCE processed and profit,    
    \item \textsc{import-only} policy, which only imports lithium minerals from foreign sources without building domestic mining infrastructure,
    \item a \textsc{deterministic} optimization approach that solves the optimization problem assuming the initial estimate of the reserves is correct,
    \item a Monte Carlo-based \textsc{stochastic} optimization approach that randomly samples $N$ scenarios from the uncertainty distribution of the reserves and solves to maximize the expected profit.
\end{itemize}

\subsection{Experimental Setup}

\begin{figure}[t]
    \centering
    \includegraphics[width=0.5\linewidth]{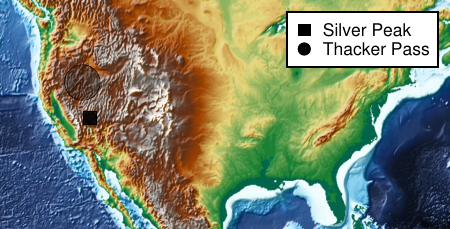}
    \caption{Domestic lithium reserves. The size of the markers represents the estimated volume of reserves at each site. The transparency of the markers represents the uncertainty of the estimates.}
    \label{fig:li_reserves_domestic}
\end{figure}

\begin{figure}
    \centering
    \includegraphics[width=0.5\linewidth]{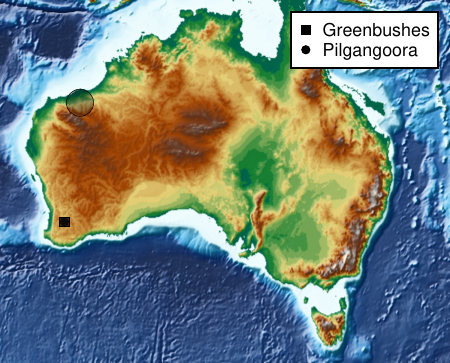}
    \caption{Foreign lithium reserves. The size of the markers represents the estimated volume of reserves at each site. The transparency of the markers represents the uncertainty of the estimates.}
    \label{fig:li_reserves_foreign}
\end{figure}

We consider a simplified model of the U.S. lithium supply chain with four candidate sites, 2 domestic and 2 foreign. Our goal is to make decisions about when and where to perform exploration while selecting where to build lithium mines. We have some initial belief about the size of each reserves and want to avoid mining domestically too early (to avoid social penalty),  minimize environmental impact, meet lithium demand, and achieve economic profitability. As an illustration, the location, reserve estimate sizes, and uncertainty of the estimates are shown in Figures~\ref{fig:li_reserves_domestic} and~\ref{fig:li_reserves_foreign}. The parameters for the model are summarized in~\autoref{tab:exp_params}.

\begin{table}[h]
    \centering
    \caption{Experimental Parameters}
    \label{tab:exp_params}
    \resizebox{0.8\columnwidth}{!}{%
    \begin{tabular}{@{}lcl@{}}
    \toprule
    Parameter & Symbol & Value \\
    \midrule
    Number of sites & $n$ & 4 (2 domestic, 2 foreign) \\
    Domestic sites & $J_d$ & $\{1, 2\}$ \\
    Foreign sites & $J_f$ & $\{3, 4\}$ \\
    LCE reserve discretization & -- & \SI{1000} Mt intervals \\
    Observation noise & $\sigma_o$ & \SI{6000} Mt \\
    Exploration cost & $c_e$ & \SI{50}[\$]M \\
    Build cost & $c_b$ & \SI{400}[\$]M \\
    Restoration cost & $c_r$ & \SI{100}[\$]M \\
    Transportation cost & $c_t$ & [\SI{0.001}[\$], \SI{0.001}[\$], \SI{0.05}[\$], \SI{0.05}[\$]] per Mt \\
    Processing cost & $c_p$ & \SI{3000}[\$] per Mt \\
    Lithium price (processed) & $p_{Li}$ & \SI{15000}[\$] per ton \\
    Discount factor & $\gamma$ & 0.97 \\
    Planning horizon & -- & 30 years \\
    Extraction factor & $\rho$ & 0.08 \\
    Domestic mining delay goal & $t_d$ & 10 years \\
    Domestic mining penalty & $p_d$ & \SI{100}[\$]M \\
    Initial reserves & $v_1, \ldots, v_4$ & \SI{100000}, \SI{50000}, \SI{100000}, \SI{50000} Mt \\
    Annual mine yield & $\phi_1, \phi_3$ & $N(5000, 50)$ Mt/year \\
     & $\phi_2, \phi_4$ & $N(2000, 50)$ Mt/year \\
    Annual transportation loss & $\psi_1, \psi_2$ & $N(1, 0.2)$ Mt/year \\
     & $\psi_3, \psi_4$ & $N(100, 10)$ Mt/year \\
    CO$_2$ emission factor & $e_1, \ldots, e_4$ & 3, 4, 6, 7 Mt/Mt \\
    Objective function weights & $w_1, \ldots, w_4$ & 0.5, 3.2, 0.5, 0.5 \\
    Demand year 1-5 & $d_1, \ldots, d_5$ & $U(50, 100)$ Mt/year \\
    Demand year 6-10 & $d_6, \ldots, d_{10}$ & $U(100, 200)$ Mt/year \\
    Demand year 11-20 & $d_{11}, \ldots, d_{20}$ & $U(200, 400)$ Mt/year \\
    Demand year 21-30 & $d_{21}, \ldots, d_{30}$ & $U(400, 600)$ Mt/year \\
    \bottomrule
    \end{tabular}
    }
\end{table}

We note that the parameters are chosen for illustrative purposes and do not represent actual values. The goal of the experiment is to demonstrate the effectiveness of the POMDP framework in optimizing lithium sourcing decisions under uncertainty and to compare the results with benchmarks. We evaluate the performance of the POMDP solver with metrics of CO$_2$ emissions, lithium yield, and economic profit, and compare the results in cases where initial beliefs are accurate and inaccurate. In the accurate case, the true lithium reserves are within 95\% confidence interval of the initial estimate. In the inaccurate case, the true lithium reserves are far outside the 95\% confidence interval.

\subsection{POMDP Solvers}

We use two POMDP solvers: POMCPOW~\citep{sunberg2018online} and DESPOT~\citep{somani2013despot}. 
\begin{itemize}
    \item \textsc{pomcpow}: an online POMDP solver that uses a rollout policy to simulate the agent's interactions with the environment and estimate the value of different actions. It is a versatile algorithm that can solve large POMDPs with high-dimensional state, action, and observation spaces through the use of progressive widening and tree pruning techniques. We use \textsc{pomcpow} to solve the lithium sourcing problem and optimize the agent's decisions under uncertainty.
    \item \textsc{despot}: an online POMDP solver that uses a deterministic sampling-based approach to approximate the value function of the POMDP, allowing it to scale to large problems with high-dimensional state spaces and long planning horizons. We use \textsc{despot} to test the scenario in which the domestic mining rate is doubled to meet the demand.
\end{itemize}

\subsection{Deterministic and Stochastic Optimization Models}

For the \textsc{deterministic} optimization approach, we solve a mixed integer linear programming (MILP) model assuming the initial estimate of the reserves is correct. Given an initial estimate of the reserves, we solve a model with binary decision variables representing for each site at each time step, which action is taken (explore, build, restore, import, or do nothing), forcing only one action to be taken at each time step. We also have a binary variable for each site at each time step, representing whether the site is operational or not, and integer variables for the amount of LCE extracted from each site (imported or domestically mined) as well as the amount of LCE processed at the plant. The objective is then to maximize the (linearized) weighted objective over the planning horizon. 

For the Monte Carlo-based \textsc{stochastic} optimization approach, we sample $N=1000$ scenarios from the uncertainty distribution of the reserves. We then solve to maximize the expected weighted objective over the scenarios.

\subsection{Results}

We present the results of our numerical experiments for the different policies in two scenarios: when the initial belief about lithium reserves is close to the ground truth (\autoref{tab:results_accurate}) and when it deviates significantly (\autoref{tab:results_inaccurate}). The metrics include the year domestic mining begins, amount of minerals processed, CO$_2$ emissions, percentage of demand unfulfilled, profit, and total discounted reward.

\begin{table}
    \centering    \caption{ Performance comparison when initial belief is close to ground truth ($\uparrow$: higher is better, $\downarrow$: lower is better; \textbf{bold} indicates best performance)}
    \label{tab:results_accurate}
    \resizebox{\columnwidth}{!}{%
    \begin{tabular}{@{}lcccccc@{}}
    \toprule
    Policy & \begin{tabular}[c]{@{}c@{}}Domestic\\Start (year)$\uparrow$\end{tabular} & \begin{tabular}[c]{@{}c@{}}Processed\\(Mt)$\uparrow$\end{tabular} & \begin{tabular}[c]{@{}c@{}}CO$_2$\\(Mt)$\downarrow$\end{tabular} & \begin{tabular}[c]{@{}c@{}}Demand\\Unfulfilled (\%)$\downarrow$\end{tabular} & \begin{tabular}[c]{@{}c@{}}Profit\\(\$B)$\uparrow$\end{tabular} & \begin{tabular}[c]{@{}c@{}} Discounted\\Reward$\uparrow$\end{tabular} \\
    \midrule
\textsc{random} & \textbf{13.2 $\pm$ 4.3} & 8,079.6 $\pm$ 421.1 & 544.0 $\pm$ 25.0 & 14.05 $\pm$ 2.5 & 8.17 $\pm$ 0.4 & 566.87 $\pm$ 28.3 \\
\textsc{import-only} & N/A & 4,477.8 $\pm$ 223.9 & 383.0 $\pm$ 19.2 & 52.36 $\pm$ 2.6 & 4.23 $\pm$ 0.2 & -940.59 $\pm$ 47.0 \\
\textsc{greedy} & 10.0 $\pm$ 0.0 & 9,345.1 $\pm$ 465.2 & 526.0 $\pm$ 26.3 & 0.61 $\pm$ 0.1 & 10.08 $\pm$ 0.5 & 1,422.04 $\pm$ 71.1 \\
\textsc{deterministic} & 10.0 $\pm$ 0.0 & 9,275.1 $\pm$ 463.8 & 229.0 $\pm$ 11.5  & 1.33 $\pm$ 0.2 & \textbf{10.34 $\pm$ 0.5} & 1,755.39 $\pm$ 87.8 \\
\textsc{stochastic} & 10.0 $\pm$ 0.0 & \textbf{9,334.0 $\pm$ 452.2} & \textbf{239.0 $\pm$ 11.0} & \textbf{0.60 $\pm$ 0.1} & 10.05 $\pm$ 0.5 & \textbf{1,762.81 $\pm$ 88.1} \\
\textsc{pomcpow} & 11.2 $\pm$ 0.5 & 9,341.3 $\pm$ 471.8 & 340.0 $\pm$ 17.0 & 0.62 $\pm$ 0.1 & 9.75 $\pm$ 0.5 & 1,632.02 $\pm$ 81.6 \\
\textsc{despot} & 11.3 $\pm$ 0.2 & 9,290.1 $\pm$ 464.5 & 368.0 $\pm$ 18.4 & 1.17 $\pm$ 0.2 & 10.24 $\pm$ 0.5 & 1,677.63 $\pm$ 98.9  \\
    \bottomrule
    \end{tabular}
    }
\end{table}

\begin{table}
    \centering    \caption{Performance comparison when initial belief is far from the ground truth ($\uparrow$: higher is better, $\downarrow$: lower is better; \textbf{bold} indicates best performance)}
    \label{tab:results_inaccurate}
    \resizebox{\columnwidth}{!}{%
    \begin{tabular}{@{}lcccccc@{}}
    \toprule
   Policy & \begin{tabular}[c]{@{}c@{}}Domestic\\Start (year)$\uparrow$\end{tabular} & \begin{tabular}[c]{@{}c@{}}Processed\\(Mt)$\uparrow$\end{tabular} & \begin{tabular}[c]{@{}c@{}}CO$_2$\\(Mt)$\downarrow$\end{tabular} & \begin{tabular}[c]{@{}c@{}}Demand\\Unfulfilled (\%)$\downarrow$\end{tabular} & \begin{tabular}[c]{@{}c@{}}Profit\\(\$B)$\uparrow$\end{tabular} & \begin{tabular}[c]{@{}c@{}} Discounted\\Reward$\uparrow$\end{tabular} \\
    \midrule
\textsc{random} & 11.2 $\pm$ 5.2 & 8,572.5 $\pm$ 428.6 & 361.0 $\pm$ 18.1 & 8.80 $\pm$ 0.5 & 9.06 $\pm$ 0.5 & 1,247.25 $\pm$ 62.4 \\
\textsc{import-only} & N/A & 4,527.8 $\pm$ 231.2 & 383.0 $\pm$ 19.8 & 51.83 $\pm$ 2.8 & 4.30 $\pm$ 0.3 & -906.73 $\pm$ 48.2 \\
\textsc{greedy} & 10.0 $\pm$ 0.0 & 9,341.3 $\pm$ 467.0 & 526.0 $\pm$ 26.3 & 0.64 $\pm$ 0.2 & 10.09 $\pm$ 0.6 & 1,422.31 $\pm$ 73.5 \\
\textsc{deterministic} & 10.0 $\pm$ 0.0 & 9,265.9 $\pm$ 458.9 & 368.0 $\pm$ 19.2 & 1.43 $\pm$ 0.2 & \textbf{10.33 $\pm$ 0.4} & 1,750.90 $\pm$ 85.3 \\
\textsc{stochastic} & 10.0 $\pm$ 0.0 & 9,342.9 $\pm$ 471.2 & 339.0 $\pm$ 17.8 & \textbf{0.64 $\pm$ 0.1} & 10.05 $\pm$ 0.5 & 1,762.75 $\pm$ 89.4 \\
\textsc{pomcpow} & \textbf{11.7 $\pm$ 0.2} & \textbf{9,344.1 $\pm$ 465.5} & 280.0 $\pm$ 16.2 & 0.64 $\pm$ 0.2 & 9.75 $\pm$ 0.6 & 1,814.21 $\pm$ 94.8 \\
\textsc{despot} & 11.2 $\pm$ 0.3 & 9,281.2 $\pm$ 469.8 & \textbf{229.0 $\pm$ 12.4} & 1.27 $\pm$ 0.2 & 10.23 $\pm$ 0.5 & \textbf{1,973.15 $\pm$ 96.5} \\
    \bottomrule
    \end{tabular}
    }
\end{table}

\begin{figure}
    \centering
    \includegraphics[width=\textwidth]{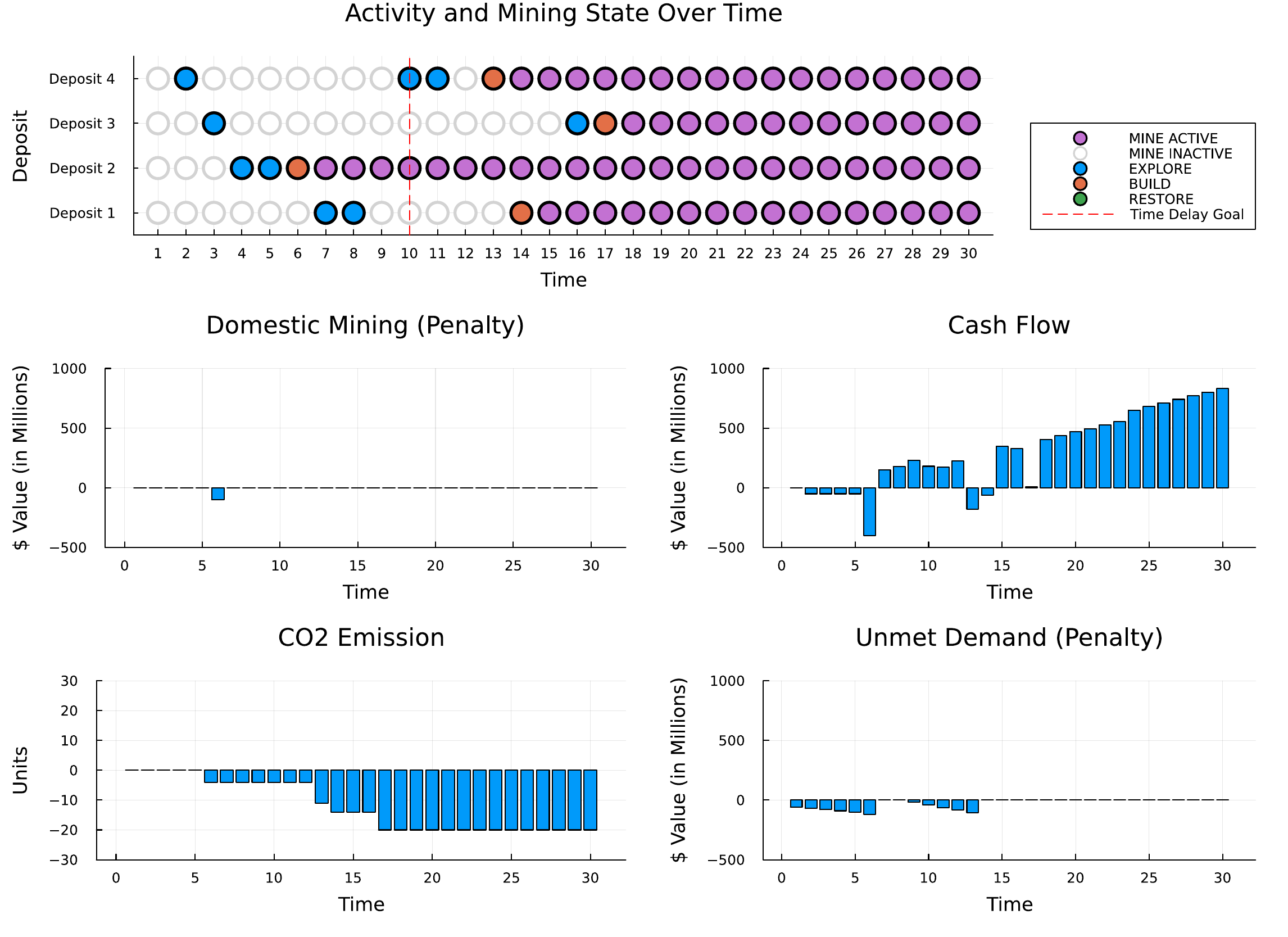}
    \caption{Results of the \textsc{random}  policy. It is clear that the policy is not optimized for any of the objectives.}
    \label{fig:results1}
\end{figure}

\begin{figure}
    \centering
    \includegraphics[width=\textwidth]{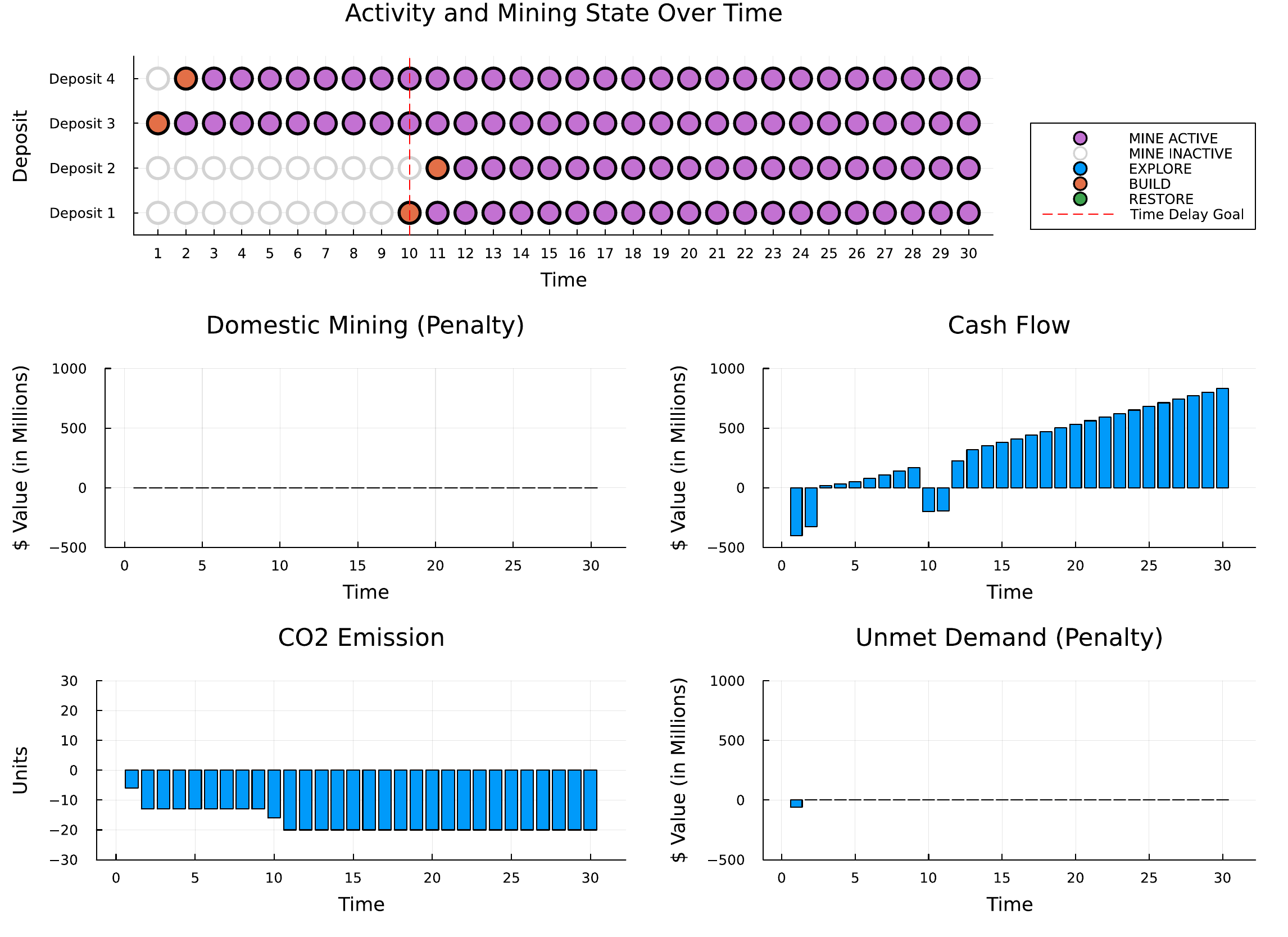}
    \caption{Results of the \textsc{greedy}  policy. The policy is optimized for the amount of LCE processed, and avoiding domestic mining penalties.}
    \label{fig:results2}
\end{figure}

\begin{figure}
    \centering
    \includegraphics[width=\textwidth]{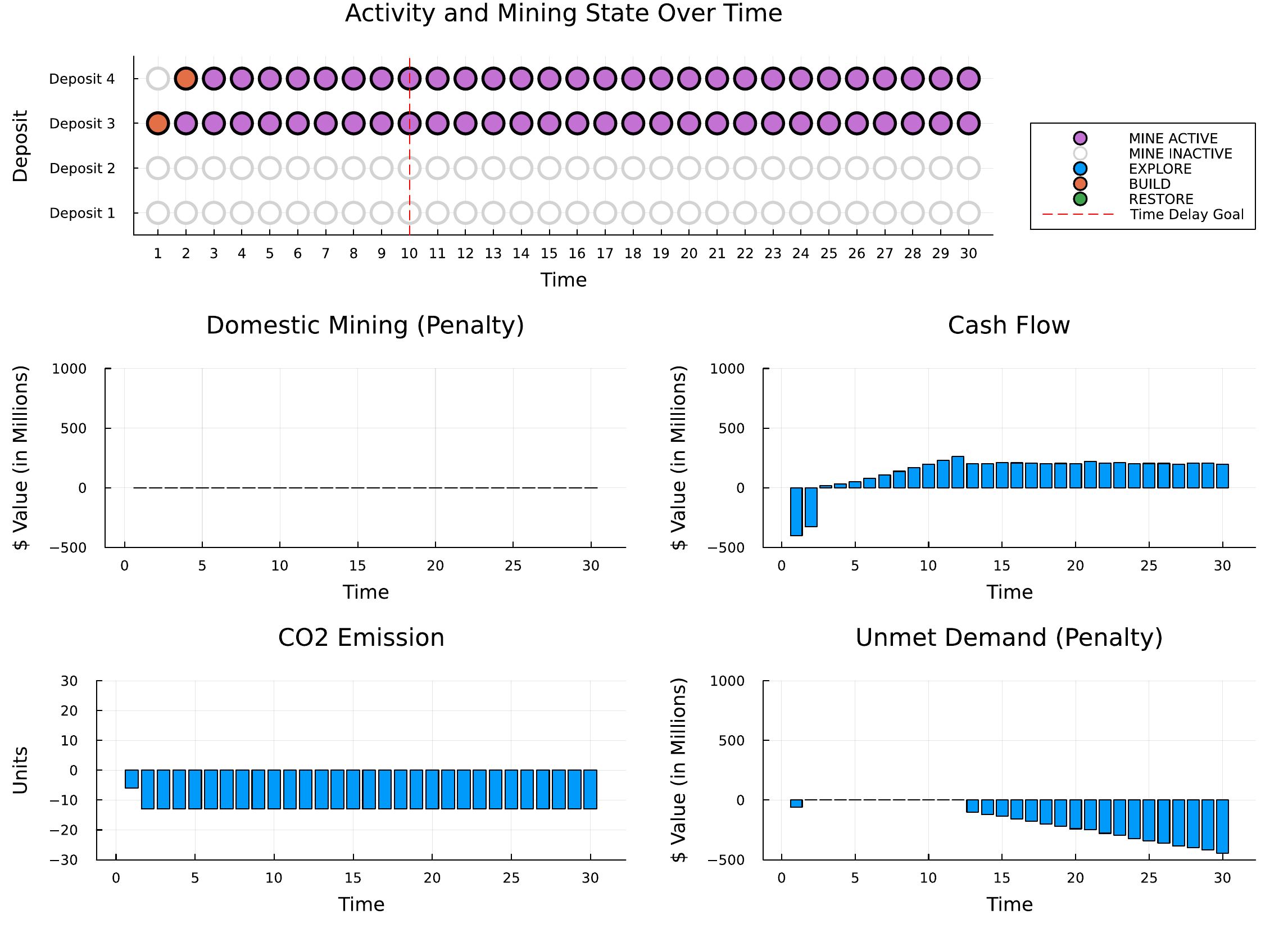}
    \caption{Results of the \textsc{import-only}  policy. The policy only imports from foreign deposits, thus failing to meet demands at later time periods.}
    \label{fig:results3}
\end{figure}

\begin{figure}
    \centering
    \includegraphics[width=\textwidth]{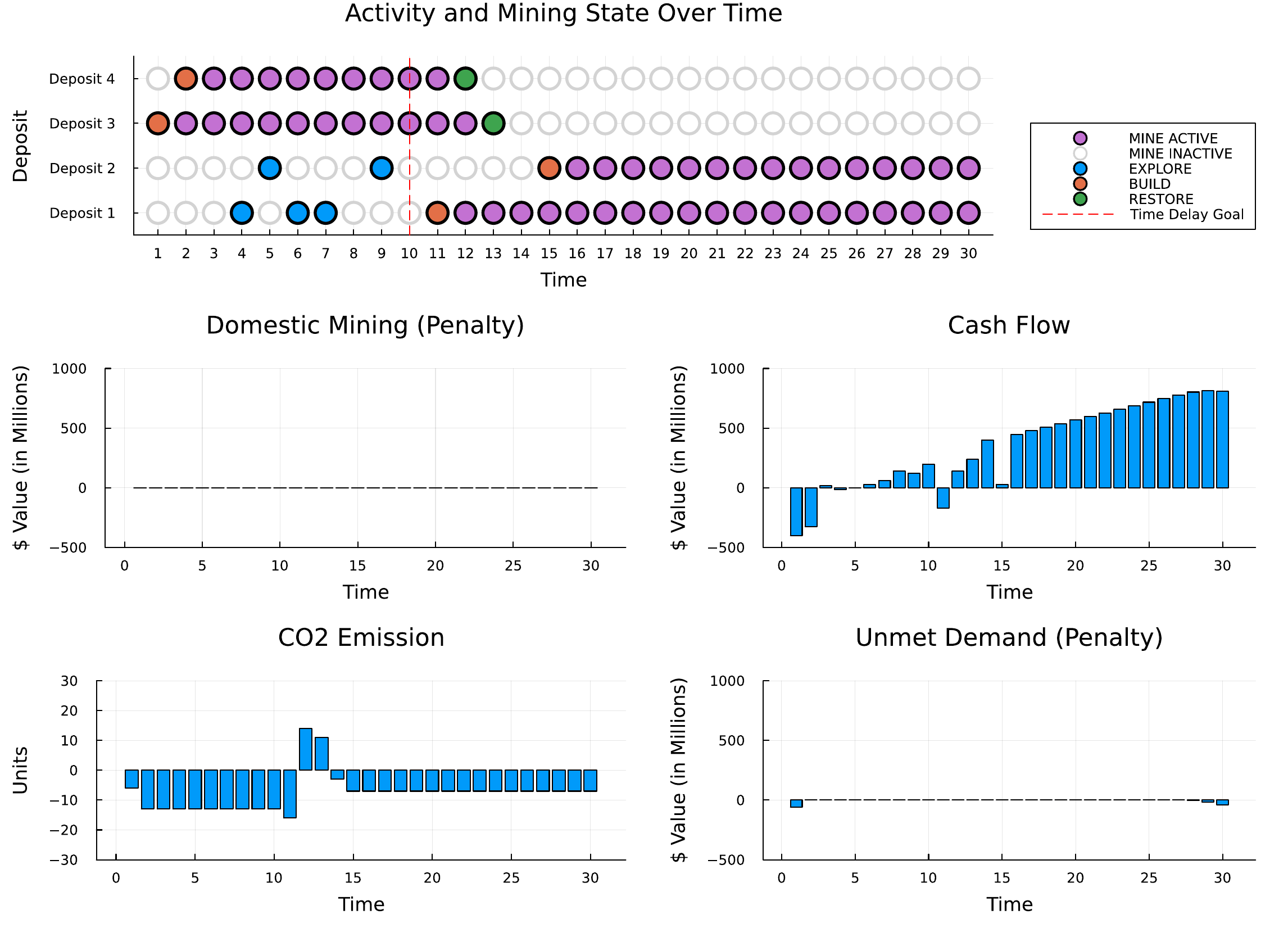}
    \caption{Results of the POMDP (\textsc{despot}) policy, succeeding in achieving a balanced objectives and thus highest discounted rewards.}
    \label{fig:results4}
\end{figure}

\begin{figure}[ht!]
    \centering
    \includegraphics[width=\textwidth]{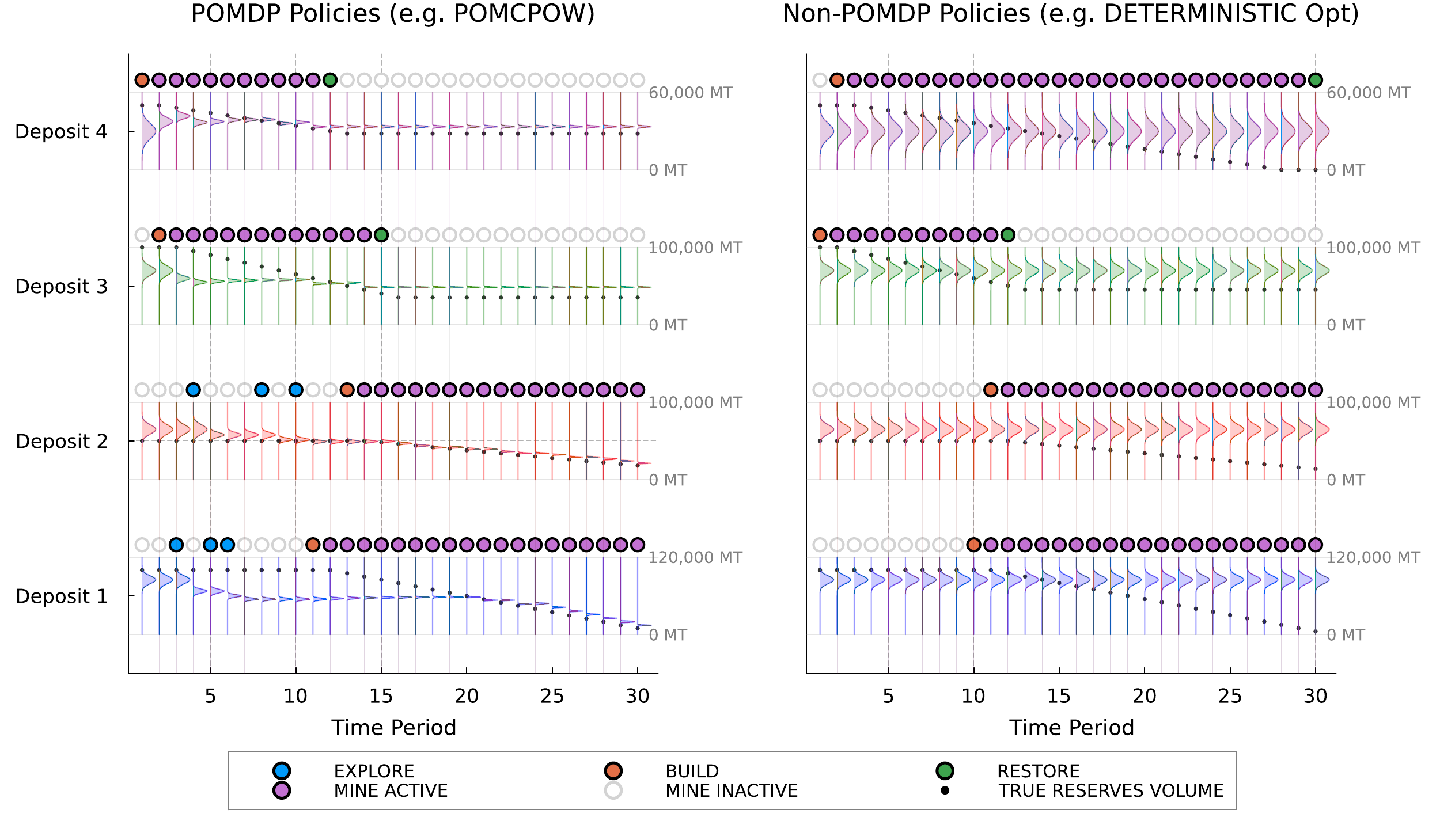}
    \caption{Comparison of Li reserve uncertainty estimates over time for POMDP and non-POMDP approaches. Even if the resulting mining plan is almost the same, the POMDP approach is able to better track the ground truth and updates its uncertainty estimates over time through belief updates, while the non-POMDP approaches is not able to do so.}
    \label{fig:results5}
\end{figure}

We also show an example of a simulation rollout for different policies in Figures~\ref{fig:results1}--\ref{fig:results4} to illustrate how the actions taken and the resulting metrics correspond to the strengths and limitations of each policy. Finally, we illustrate the strength of POMDP belief updating especially when dealing with inaccurate initial reserve estimates in~\autoref{fig:results5}.

\section{Discussion}
\label{sec:discussion}

Our analysis of different lithium sourcing policies reveals important insights about the tradeoffs between domestic production, environmental impact, and supply chain resilience, particularly highlighting the crucial role of addressing geological uncertainty in strategic decision-making. 

\subsection{Benchmarking Policies}

The experimental results demonstrate a stark contrast in policy performance based on the accuracy of initial geological estimates. When initial beliefs about reserves are accurate (\autoref{tab:results_accurate}), the \textsc{deterministic} optimization achieves strong performance with \SI{10.34}[\$]B in profits and only 1.33\% unfulfilled demand. The \textsc{stochastic} approach performs similarly well, with \SI{10.05}[\$]B in profits and 0.60\% unfulfilled demand. This suggests that with accurate information, traditional optimization approaches can effectively balance competing objectives.

However, in the more realistic scenario where initial geological estimates are inaccurate (\autoref{tab:results_inaccurate}), the POMDP-based approaches demonstrate superior robustness. While the \textsc{deterministic} and \textsc{stochastic} approaches maintain similar performance levels, the \textsc{despot} algorithm achieves the best overall performance with \SI{10.23}[\$]B in profits, 1.27\% unfulfilled demand, and notably lower CO$_2$ emissions at 229.0 Mt, followed by \textsc{pomcpow} with competing results, compared to other approaches. This robustness stems from the POMDP's ability to systematically explore and update beliefs about reserves, as illustrated in ~\autoref{fig:results5}, where POMDP belief updates steer the estimated reserves toward the true value and incorporate them throughout the planning process, while the traditional non-POMDP approaches simply do not have such a mechanism. Given the high standard deviation in the results however, further statistical analysis would be needed to establish the significance of these performance differences.

The \textsc{import-only} policy performs poorly across both scenarios, with over 50\% of demand unfulfilled and profits of only \SI{4.23}[\$]B--\SI{4.30}[\$]B, highlighting the importance of developing domestic production capacity. The \textsc{random} policy, while achieving surprisingly decent profits (\SI{8.17}[\$]B--\SI{9.06}[\$]B), shows high variability in the year that domestic mining starts (standard deviation of 4.3--5.2 years) and suboptimal CO$_2$ emissions (see~\autoref{fig:results2}).

\subsection{Achieving Critical Minerals Self-Sufficiency}

Our results highlight the importance of balanced investment in domestic mining capacity. The POMDP policies (\textsc{despot} and \textsc{pomcpow}) suggest starting domestic mining around year 11 (see~\autoref{fig:results4}), slightly later than the other approaches which start mostly in year 10 (see~\autoref{fig:results2}). This slight delay allows for better information gathering while still maintaining high processing volumes (around 9,300 Mt) and low unfulfilled demand ($\leq1.27$\%).

The \textsc{import-only} policy's poor performance (processing only about 4,500 Mt) demonstrates the risks of overreliance on foreign sources (see~\autoref{fig:results3}). However, the success of POMDP approaches in achieving low CO$_2$ emissions (229-340 Mt compared to 526 Mt for \textsc{greedy}) shows that domestic production can be environmentally responsible when properly optimized.

\subsection{Policy Implications}

The results highlight several critical considerations for the US lithium supply chain strategy. Most of the policies initiate foreign mining right away to fulfill immediate demands (Figure~\ref{fig:results2}-\ref{fig:results4}). Soon after that, POMDP policies (\textsc{pomcpow} and \textsc{despot}) start advocating for domestic explorations, suggesting that early investment in the form of exploration in potential sites to obtain better reserve estimates may be optimal in the long run despite higher initial costs (see ~\autoref{fig:results5}). This contrasts with the current U.S. approach of heavy reliance on imports with minimal domestic exploration. POMDP policies also suggest that the U.S. does not necessarily need to build domestic mining immediately when the infrastructure is ready (at year 10), but could still delay a little bit if foreign imports can still satisfy demands (and start at years 11 and 15). However, the key insight is that infrastructure and accurate reserve estimates (from early investments in R\&D and explorations) should be in place to allow rapid domestic mining establishments when needed.

We also note that the mining rates we set in this study might be a bit higher than what is currently available in practice. This mining rate scenario indicates that investment in mining technology and efficiency is crucial for achieving supply chain independence for critical minerals in the U.S. The \textsc{import-only} policy, while showing lower immediate costs, exposes vulnerabilities in supply chain security in the long run. Furthermore, both optimized POMDP policies still require domestic lithium projects later in the years. Therefore, similar projects, like Thacker Pass, where permitting delays have impacted development timelines, require careful attention and collaboration between environmentalists, policymakers, and key industry players.

It is also important to note that while our simulation study results support domestic production expansion, they also indicate the continued importance of maintaining diverse supply sources by strengthening partnerships with allies like Australia, investing in joint technology development initiatives, and more importantly, proactive collaborations among key stakeholders to ensure long-term sustainability. Policies that support sustainability aims—such as carbon pricing or incentives to encourage cleaner extraction methods, investment in renewable energy for mining operations, and development of recycling infrastructure to reduce primary production needs (see e.g.,~\citep{kushnir2012time})—still play a major role in offsetting mining carbon footprints.

Finally, we note that our research offers only a starting point for ways to integrate geological uncertainty in the strategic decision-making of critical mineral supply chains. We only use synthetic data to illustrate the potential use cases. Future research could explore more practical policymaking approaches supported with more realistic datasets. Also, more granular policy scenarios, including varied carbon pricing schemes, different technological advancement trajectories, and alternative international partnership configurations, should be explored. Additionally, extending the analysis to consider potential disruptions from climate change impacts or geopolitical events could provide valuable insights for long-term policy planning. The framework presented here provides a foundation for such analyses, offering a quantitative approach to evaluating the complex tradeoffs inherent in critical mineral supply chain policy.

\section{Conclusion}\label{sec:conclusion}

We present a POMDP framework for optimizing lithium sourcing decisions under uncertainty, with particular emphasis on managing geological uncertainty in strategic supply chain planning. Our numerical experiments demonstrate that while deterministic optimization can achieve superior performance when initial geological estimates are accurate, its performance deteriorates dramatically when these estimates deviate from reality. In contrast, POMDP-based approaches maintain robust performance even with inaccurate initial beliefs through a belief update mechanism. These results highlight three key findings: First, the ability to actively reduce uncertainty through systematic exploration, as enabled by the POMDP framework, is crucial for robust long-term performance in mineral supply chain optimization. Second, while stochastic optimization is an improvement over deterministic approaches under uncertainty, it falls short of POMDP performance, suggesting that scenario sampling alone is insufficient for optimal decision-making. Third, the framework demonstrates that early investment in domestic mining capacity, guided by systematic exploration and uncertainty reduction, can effectively balance economic, environmental, and supply security objectives. Future work will explore more sophisticated POMDP formulations that incorporate additional sources of uncertainty, such as market dynamics and geopolitical factors. Additionally, the integration of advanced mining technologies and their impact on extraction rates could provide valuable insights for technology investment decisions.

\bibliographystyle{elsarticle-num}
\bibliography{ref}

\end{document}